\documentclass[10pt,twocolumn,letterpaper]{article}

\usepackage[pagenumbers]{cvpr}      

\usepackage{graphicx}
\usepackage{amsmath}
\usepackage{amssymb}
\usepackage{booktabs}

\usepackage{multirow}

\usepackage[pagebackref,breaklinks,colorlinks]{hyperref}

\usepackage[capitalize]{cleveref}
\crefname{section}{Sec.}{Secs.}
\Crefname{section}{Section}{Sections}
\Crefname{table}{Table}{Tables}
\crefname{table}{Tab.}{Tabs.}

\def\cvprPaperID{430} 
\def\confName{CVPR}
\def\confYear{2023}

\begin{document}

\title{EditableNeRF: Editing Topologically Varying Neural Radiance Fields\\ by Key Points}

\author{Chengwei~Zheng$\qquad\qquad$Wenbin~Lin$\qquad\qquad$Feng~Xu\\
School of Software and BNRist, Tsinghua University\\
}

\twocolumn[{%
\renewcommand\twocolumn[1][]{#1}%
\maketitle
\begin{center}
    \vspace{-0.4em}
    \centering
    \captionsetup{type=figure}
    \includegraphics[width=1.0\textwidth]{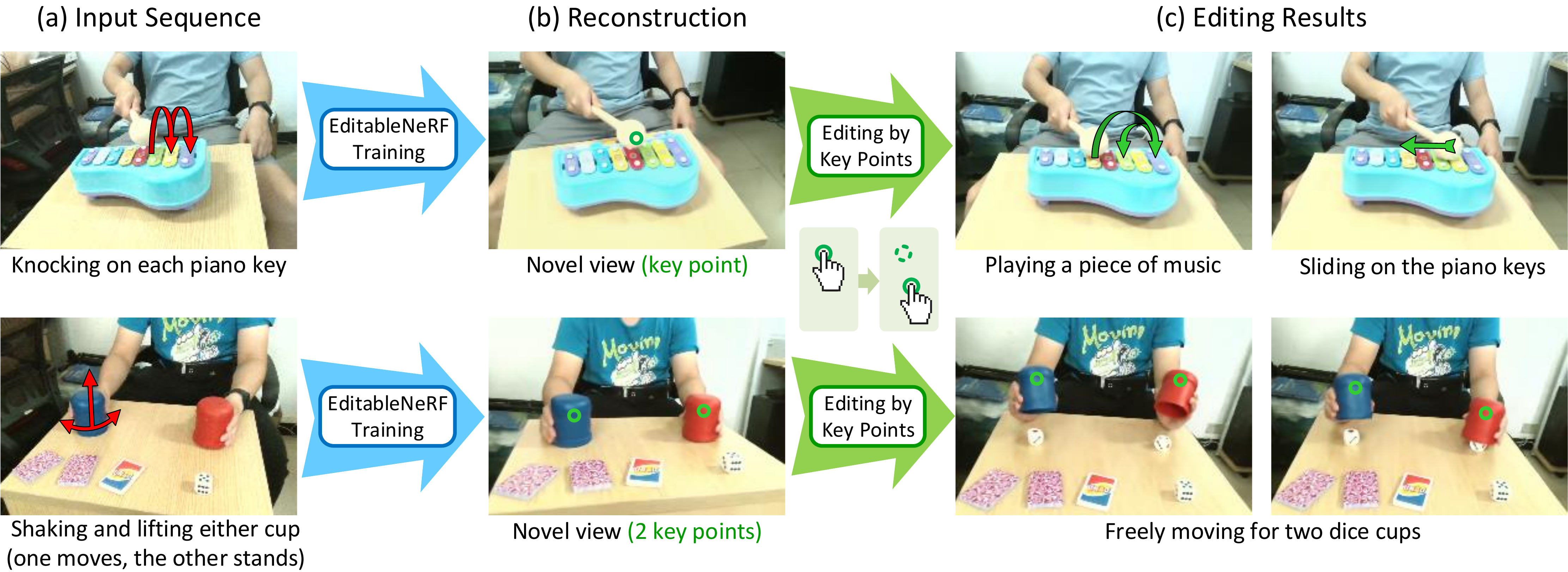}
    \captionof{figure}{Taking an image sequence (a) as input, EditableNeRF is trained fully automatically to reconstruct the captured scene (b) and can handle topological changes. After training, end-users are able to edit the scene (c) by controlling the automatically picked-out key points (circled in green in (b)). 
    Our method enables multi-dimensional editing and can generate novel scenes that are unseen during training.
    }
	\label{fig:teaser}
\end{center}%
}]

\begin{abstract}
Neural radiance fields (NeRF) achieve highly photo-realistic novel-view synthesis, but it's a challenging problem to edit the scenes modeled by NeRF-based methods, especially for dynamic scenes. We propose editable neural radiance fields that enable end-users to easily edit dynamic scenes and even support topological changes. Input with an image sequence from a single camera, our network is trained fully automatically and models topologically varying dynamics using our picked-out surface key points. Then end-users can edit the scene by easily dragging the key points to desired new positions. To achieve this, we propose a scene analysis method to detect and initialize key points by considering the dynamics in the scene, and a weighted key points strategy to model topologically varying dynamics by joint key points and weights optimization. Our method supports intuitive multi-dimensional (up to 3D) editing and can generate novel scenes that are unseen in the input sequence. Experiments demonstrate that our method achieves high-quality editing on various dynamic scenes and outperforms the state-of-the-art. Our code and captured data are available at \url{https://chengwei-zheng.github.io/EditableNeRF/}.
\end{abstract}



\section{Introduction}
Neural radiance fields (NeRF) \cite{mildenhall2020nerf} have shown great power in novel-view synthesis and enable many applications as this method achieves photo-realistic rendering \cite{gao2022nerf}. Recent techniques have further improved NeRF by extending it to handle dynamic scenes \cite{park2021nerfies, pumarola2021d, tretschk2021non} and even topologically varying scenes \cite{park2021hypernerf}. However, these works mainly focus on reconstruction itself but do not consider scene editing. Thus, for rendering, only the camera views can be changed, while the modeled scenes cannot be edited.

Recently, some frameworks have been proposed to make neural radiance fields editable in different aspects. Some of them aim to edit the reconstructed appearance and enable relighting \cite{zhang2021nerfactor, boss2021nerd, srinivasan2021nerv}; some allow controlling the shapes and colors of objects from a specific category \cite{liu2021editing, xie2021fig, jang2021codenerf, wei2022self}; and some divide the scene into different parts and the location of each part can be modified \cite{yang2021learning, zhang2021editable, yu2021unsupervised}. However, the dynamics of moving objects cannot be edited by the previous methods. And this task becomes much more challenging when the dynamics contain topological changes. Topological changes can lead to motion discontinuities (e.g., between the hammer and the piano keys, between the cups and the table in Fig. \ref{fig:teaser}) in 3D space and further cause noticeable artifacts if they are not modeled well. A state-of-the-art framework CoNeRF~\cite{kania2022conerf} tries to resolve this problem by using manual supervision. However, it only supports limited and one-dimensional editing for each scene part, requiring user annotations as supervision.

We propose EditableNeRF, editable topologically varying neural radiance fields that are trained without manual supervision and support intuitive multi-dimensional (up to three-dimensional) editing. 
The key of our method is to represent motions and topological changes by the movements of some sparse surface key points. 
Each key point is able to control the topologically varying dynamics of a moving part, as well as other effects like shadow and reflection changes through the neural radiance fields. 
This key-point-based method enables end-users to edit the scene by easily dragging the key points to their desired new positions. 

To achieve this, we first apply a scene analysis method to detect key points in the canonical space and track them in the full sequence for key point initialization. 
We introduce a network to estimate spatially-varying weights for all scene points and use the weighted key points to model the dynamics in the scene, including topological changes. 
In the training stage, our network is trained to reconstruct the scene using the supervision from the input image sequence, and the key point positions are also optimized by taking motion (optical flow) and geometry (depth maps) constraints as additional supervision. 
After training, the scene can be edited by controlling the key points' positions, and novel scenes that are unseen during training can also be generated.

The contribution of this paper lies in the following aspects:
\begin{itemize}
\setlength{\itemsep}{0pt}
\setlength{\parskip}{0pt}
\setlength{\parsep}{0pt}
  \item Key-point-driven neural radiance fields achieving intuitive multi-dimensional editing even with topological changes, without requiring annotated training data.
  \item A weighted key points strategy modeling topologically varying dynamics by joint key points and weights optimization.
  \item A scene analysis method to detect and initialize key points by considering the dynamics in the scene.
\end{itemize}

  
\section{Related Work}

\subsection{Novel-View Synthesis}

Many methods achieve rendering novel-view images by reconstructing scenes and objects into meshes\cite{guo2017real, dou2017motion2fusion, du2018montage4d, thies2019deferred, zheng2021dtexfusion}, neural voxels \cite{lombardi2019neural, sitzmann2019deepvoxels}, and multi-plane images \cite{zhou2018stereo, flynn2019deepview, mildenhall2019local}. Besides these methods based on discrete representations, some methods also achieve novel-view synthesis by using continuous representations \cite{sitzmann2019scene, park2019deepsdf} and have shown great potential in this task.

Neural radiance fields (NeRF)~\cite{mildenhall2020nerf} achieve photo-realistic rendering in novel-view synthesis by leveraging continuous implicit functions of density and view-dependent color to represent static scenes. To handle dynamic scenes, time-variant latent codes could be used to encode time-variant components based on NeRF, but requiring multi-view video inputs \cite{li2022neural}. To further enable dynamic reconstructions from a single-view sequence, deformation fields implemented by MLPs are applied to warp objects in each frame into a canonical space \cite{park2021nerfies, pumarola2021d, tretschk2021non}. Some methods also utilize estimated depth maps \cite{xian2021space}, ToF depth images \cite{attal2021torf}, or optical scene flow \cite{li2021neural} to improve the performance of dynamic neural radiance fields. HyperNeRF \cite{park2021hypernerf} further extends dynamic NeRF to reconstruct topologically varying scenes by modeling canonical spaces with different topology states into a unified continuous hyperspace, and the discontinuous deformations in 3D space caused by topological changes can be modeled by continuous functions in hyperspace. However, unlike traditional explicit representations such as triangular meshes, NeRF-based methods represent the scenes by implicit functions, making the modeled scenes difficult to be edited.

\subsection{Editing on Neural Radiance Fields}

As our method focuses on NeRF editing, we mainly discuss NeRF-based methods that support user editing in this section. For editing on explicit representations or other implicit representations, please refer to  \cite{yuan2021revisit, deng2021deformed, zheng2021deep, sitzmann2020implicit}.

One approach to edit neural radiance fields is to segment the scene into different components and build MLP for each component \cite{yang2021learning, zhang2021editable, yu2021unsupervised}. Assuming that different components are individual, this representation allows control of the placements and the relative positions of these components, as well as deleting or reduplicating a component. But these methods do not support editing the dynamics inside a component and only result in limited applications.

\begin{figure*}[t]
    \vspace{-1.5em}
	\centerline{\includegraphics[width=1.0\linewidth]{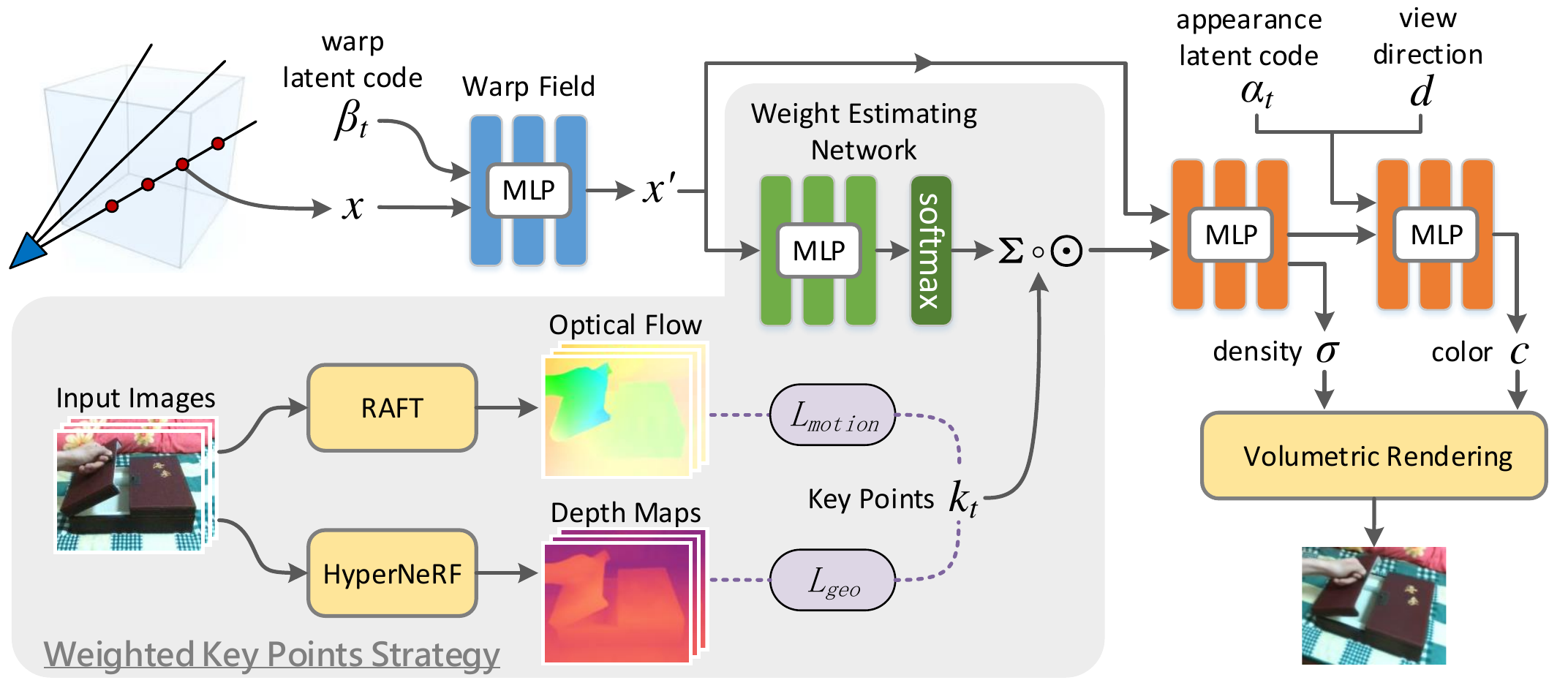}}
	\caption{EditableNeRF pipeline. The query point $x$ is first warped into the canonical space by a warp field and a latent code $\beta_t$ in frame $t$. Next, we compute the key point weights of this canonical point $x'$ and use it to calculate a linear combination of all key point positions $k_t$, called \emph{weighted key points}. After that, we feed the following NeRF MLP with the weighted key points and $x'$, then the output density and color are used for volumetric rendering. In the training stage, optical flow and depth maps are used to supervise key point positions.}
	\label{fig:pipeline}
    \vspace{-0.5em}
\end{figure*}

In addition, some methods achieve relighting and material editing on neural fields \cite{zhang2021nerfactor, boss2021nerd, srinivasan2021nerv} by decomposing the scene into surface normals, lights, albedo, and material. Texture editing can also be accomplished by a 3D-to-2D texture mapping \cite{xiang2021neutex}.

Besides, there are also some methods that focus on modeling a specific category of objects \cite{liu2021editing, xie2021fig, jang2021codenerf, wei2022self} instead of general objects. A common solution for this problem is to model the category of objects with conditional NeRF and use latent codes as conditions to encode the variations of different objects in this category. Then the shape and appearance can be edited by changing the latent codes or by fine-tuning the network \cite{liu2021editing}, and even controlled by text~\cite{wang2022clip} with the help of a multi-modal model. And many methods also focus on modeling editable human bodies or human faces based on NeRF representation. By using human body parametric models and skinning techniques such as SMPL~\cite{loper2015smpl}, neural radiance fields have been extended to model the human body and can be animated by controlling skeleton poses \cite{peng2021animatable, chen2021animatable, noguchi2021neural, su2021nerf, liu2021neural}. Human face parametric models also contribute to extending NeRF for human face modeling and controlling \cite{gafni2021dynamic, sun2022fenerf, hong2022headnerf, zheng2022avatar, grassal2022neural, wang2021one}, and even driven by audio \cite{guo2021ad}. However, general objects cannot be handled by these methods.

Recently, NeRF-editing~\cite{yuan2022nerf} proposes to deform NeRF on static objects by extracting explicit meshes, deforming the meshes, and transferring the deformations back into the implicit representations. However, this method cannot handle dynamic scenes. CoNeRF~\cite{kania2022conerf} proposes an attribute re-rendering method based on dynamic NeRF. This method requires users to provide annotations in several frames, including masks for every dynamic part and their corresponding attribute values, for network training. Then these parts could be edited by controlling the one-dimensional attribute values. We will show our advantages against CoNeRF in Sec. \ref{sec:comparisons}.


\section{EditableNeRF}

Input with color image sequence, our method can reconstruct the captured scene fully automatically based on neural radiance fields, and the topologically varying dynamics are modeled using surface key points. After reconstruction, end-users can edit the scene by controlling the key points.

Our pipeline is shown in Fig. \ref{fig:pipeline}. First, We use two methods, HyperNeRF~\cite{park2021hypernerf} and RAFT~\cite{teed2020raft}, to derive the depth maps of input frames and optical flow between adjacent input images, respectively. Then we apply a scene analysis method to detect and initialize key points for each frame (Sec. \ref{sec:initialization}). After that, our NeRF-based network (Sec. \ref{sec:network}) can be trained fully automatically (Sec. \ref{sec:training}) to model the captured scene based on our weighted key points strategy. When the reconstruction is finished, the reconstructed scene can be edited by dragging the key points to desired positions (Sec. \ref{sec:editing}).

\subsection{Network}
\label{sec:network}
We first introduce our network architecture, which is shown in Fig. \ref{fig:pipeline}. Our network represents the scene as a field of density and radiance~\cite{mildenhall2020nerf}. Given a query point, similarly to other dynamic NeRF methods~\cite{park2021nerfies, park2021hypernerf}, we first use a warp field to model slight movements:
\begin{equation}
    \label{eq:wrap}
    x'=T(x, \beta_t).
\end{equation}
Here the warp filed $T$ maps a query 3D point $x$ to its canonical location $x'$, and $\beta_t$ is the warp latent code in frame $t$. This warp field ensures that the scene in different frames is aligned despite some errors in input camera parameters, by using slight movements. While as discussed in \cite{park2021hypernerf}, it's hard for this continuous warp field to model discontinuous movements caused by topological changes.

\begin{figure}[t]
	\centerline{\includegraphics[width=1.0\linewidth]{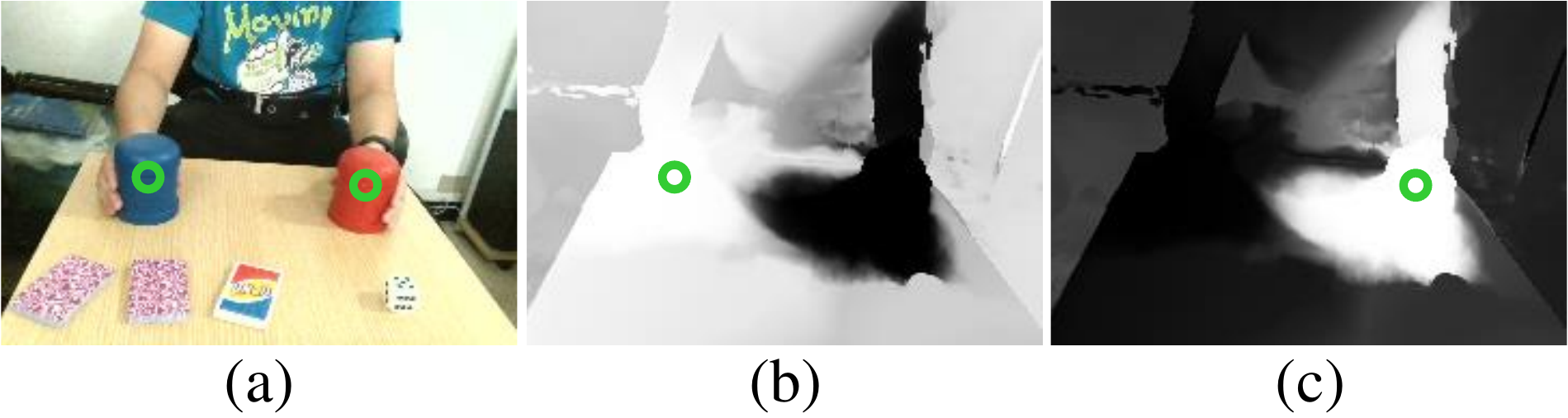}}
	\caption{Examples of key points and key point weights. (a) shows an input frame and its corresponding key points (circled in green). (b) and (c) demonstrate the weights of the two key points, respectively. These weights are obtained by using the surface points corresponding to the pixels as query points.}
	\label{fig:keypoints}
\end{figure}

Then we need to model different topology and motion states in the canonical space. We find that motions and topological changes are always related to some movements of surface points, so we achieve this modeling by making use of sparse surface key points. These 3D key points are attached to the objects' surfaces and also move with the objects. Each key point is able to control the topologically varying dynamics of a moving part and also some effects like shadow and reflection changes. An example of key points is shown in (a) of Fig. \ref{fig:keypoints}. For each moving part in the scene, we automatically select one corresponding key point, which will be detailed in Sec. \ref{sec:initialization}, and the number of key points is denoted as $N$. The key points' positions in each input frame will be optimized automatically in our training stage to achieve this modeling.

We assume that different locations in the canonical space are affected by different key points. So for a query point $x'$, an MLP followed by softmax is used to decide which key point should control its dynamics. We call this network \emph{weight estimating network}, which takes canonical coordinate $x'$ as input and outputs a weight vector $w \in \mathbb{R}^N$, indicating how each key point affects the query point $x'$. 
\begin{equation}
    \label{eq:mask}
    w=W(x').
\end{equation}
An example of these spatially-varying key point weights is shown in (b) and (c) of Fig. \ref{fig:keypoints}. 

We then construct a \emph{weighted key points} vector $p$ by taking a linear combination of all key point positions $k$ to model the topologically varying dynamics at $x'$.
\begin{equation}
    \label{eq:mul_kpoint}
    p_t(x')=\sum_{i=1}^N w^i(x') \cdot k_t^i.
\end{equation}
The superscript $i$ is the index of key points, and the subscript $t$ is the frame index. If there is only one object that moves and causes topological changes, our method will model this scene with only one key point ($N = 1$), and (\ref{eq:mul_kpoint}) becomes $p_t(x') = k_t$ because the softmax always outputs a scalar $1$. So we directly set $p_t(x')$ to be $k_t$ in this situation. 

Next, the 3D canonical coordinate $x'$ and the weighted key points $p$ are concatenated to construct a coordinate in hyperspace for topologically varying scene modeling. This hyperspace is proposed in HyperNeRF~\cite{park2021hypernerf}. In addition to 3D space, HyperNeRF makes use of ambient dimensions to model objects in hyperspace, and different topology states are encoded with different ambient coordinates. Discontinuous deformations caused by topological changes in 3D space can be modeled by continuous functions (such as MLP) in hyperspace, and more details can be found in \cite{park2021hypernerf}. Here we use the weighted key points $p$ as ambient coordinates, modeling topologically varying dynamics.

Finally, the following NeRF MLP is fed with this 6D coordinate in hyperspace:
\begin{equation}
    \label{eq:mul_nerf}
    (c, \sigma)=H(x' \oplus  p_t(x'), d, \alpha_t).
\end{equation}
Here $d$ is the view direction, and $\alpha_t$ is the appearance latent code as in ~\cite{park2021hypernerf}. This NeRF MLP outputs the color $c$ and the density $\sigma$ that can be used in volumetric rendering. To render an image, we should trace the camera rays of all pixels, sample points along these rays, obtain their colors and densities, and run the volumetric rendering, which are the same as in the original NeRF~\cite{mildenhall2020nerf}.

\subsection{Loss Functions and Training}
\label{sec:training}

All the latent codes and MLP parameters are optimized in the training stage to model the scene. As we use key points to encode topologically varying dynamics, we need to additionally optimize key point positions in each input frame. To keep our key points on the object surfaces and to be time-consistent, novel losses are added in our training stage. 

First, we propose a motion loss, which constrains that the key point positions in two adjacent frames should be consistent with the optical flow from pre-trained RAFT~\cite{teed2020raft}.
\begin{equation}
    \label{eq:flow_loss}
    L_{motion}(t, i) \!=\!
    \left\| \Pi_{t+1} (k_{t+1}^i) \!-\! \Pi_t (k_t^i) \!-\! F^{t+1}_t(\Pi_t (k_t^i))\right\| ^2,
\end{equation}
where $\Pi_t$ is the projection function using the  camera pose of frame $t$, and $F^{t+1}_t$ is the optical flow from frame $t$ to frame $t+1$. This loss ensures that the 2D key point positions in different frames correspond to the same surface point.

The motion loss provides good supervision in 2D image space, while the key points are in the 3D space. Thus, a geometry loss can help to keep the key points on the object surfaces.
\begin{equation}
    \label{eq:depth_loss}
    L_{geo}(t, i)=
    \left\| \Phi_t (k_t^i) - D_t(\Pi_t (k_t^i)) \right\| ^2.
\end{equation}
Here the function $\Phi_t(k_t^i)$ calculates the distance from the key point $k_t^i$ to the camera position in frame $t$, and $D_t$ denotes the depth map rendered from HyperNeRF~\cite{park2021hypernerf} in the original camera view. This HyperNeRF is pre-trained before our training stage and takes the same input as ours. 

Besides, we also apply a reconstruction loss between the rendered RGB images $C$ and the input images $\widetilde{C}$, as well as a warp regularization loss.
\begin{equation}
    \label{eq:recon_loss}
    L_{rec}(t) = \left\| C_t(k_t, \alpha_t, \beta_t) - \widetilde{C}_t\right\|^2,
\end{equation}\begin{equation}
    \label{eq:warp_reg_loss}
    L_{reg}(t) = \frac{1}{\left\| S_t \right\|} \sum_{x \in S_t}\left\| x - T(x, \beta_t)\right\|^2,
\end{equation}
where $S_t$ is the set of surface points in frame $t$. This warp regularization loss makes sure that the warp field only models slight movements to compensate for the errors in the input camera parameters and distortions in the input images. Thus, it can avoid the ambiguity between the warp field and our weighted key points model.

\subsection{Key Point Detection and Initialization}
\label{sec:initialization}

\begin{figure}[t]
    \vspace{-0.5em}
	\centerline{\includegraphics[width=1.0\linewidth]{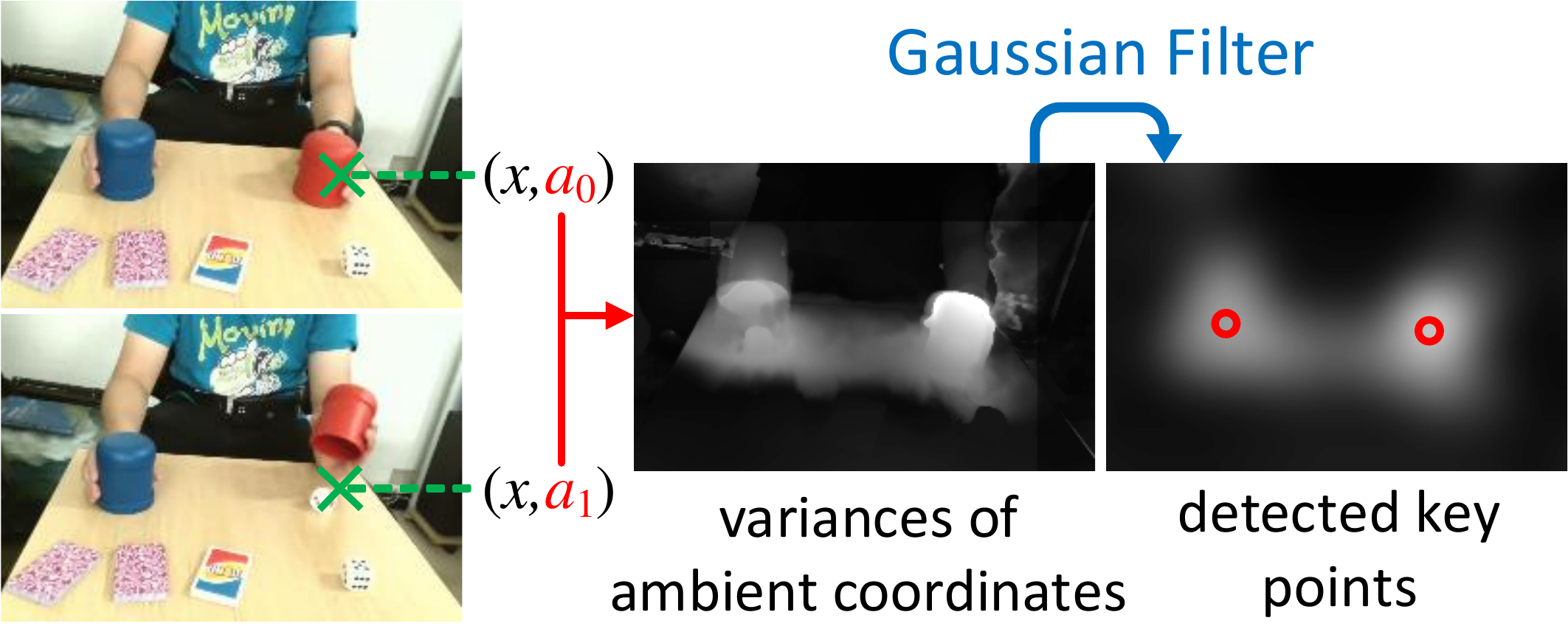}}
	\caption{A 2D visualization of our key point detection method. For each point $x$, we compute the variance of its ambient coordinates $a$ in the full sequence. Then the points with local maximum variances after a 2D Gaussian filter will be selected as our reference key points.}
	\label{fig:kp_detect}
    \vspace{-0.5em}
\end{figure}

To initialize the network training, we need to determine the key point number $N$ and obtain the initial 3D locations of key points. To achieve this, we apply a scene analysis method, which first finds reference key points in canonical space and reference frames where these reference key points are on the object surfaces, then initializes key points' positions in each frame.

As our key points are used to model topologically varying dynamics, we find the 3D points in the canonical space with dramatic dynamics including topological changes as our reference key points. Recall that HyperNeRF~\cite{park2021hypernerf} uses different ambient coordinates $a$ to encode different topology and motion states at a position. And we have already trained a HyperNeRF for depth maps in (\ref{eq:depth_loss}). So we can detect key points by making use of the ambient dimensions $a$ from this pre-trained HyperNeRF. For a point $x$, great varying of its $a$ in different frames indicates great dynamics at $x$, so we use all the positions with locally greatest variations of ambient coordinates as our reference key points. 

To be specific, for each input frame, we trace the original camera rays of all pixels and find the corresponding surface points. A 3D voxel volume in the canonical space is then built to record the ambient coordinates of these surface points. After traversing the whole input sequence, we compute the variance of ambient coordinates for every voxel, followed by a 3D Gaussian filter. And the center points of the voxels with local maximum variances after Gaussian blur will be selected as the reference key points $k_{ref}$. As our 3D version is difficult to visualize, we show a 2D version of this process in Fig. \ref{fig:kp_detect} by rendering all frames in a fixed camera view and computing the variance for each pixel, followed by a 2D Gaussian filter. 

Then for each reference key point, we need to select a reference frame in which the reference key point is on the object surface. We use a similar formulation with the geometry loss in (\ref{eq:depth_loss}) to decide this. (Here we omit the index of key points as they are handled individually.)
\begin{equation}
    \label{eq:ref_frame}
    \left\| \Phi_t (k_{ref}) - D_t(\Pi_t (k_{ref})) \right\| ^2 < \delta,
\end{equation}
where $\delta$ is a pre-defined threshold. The first frame $t$ that satisfies (\ref{eq:ref_frame}) will be selected as the reference frame $t_{ref}$. 

\begin{figure}[t]
    \vspace{-0.5em}
	\centerline{\includegraphics[width=1.0\linewidth]{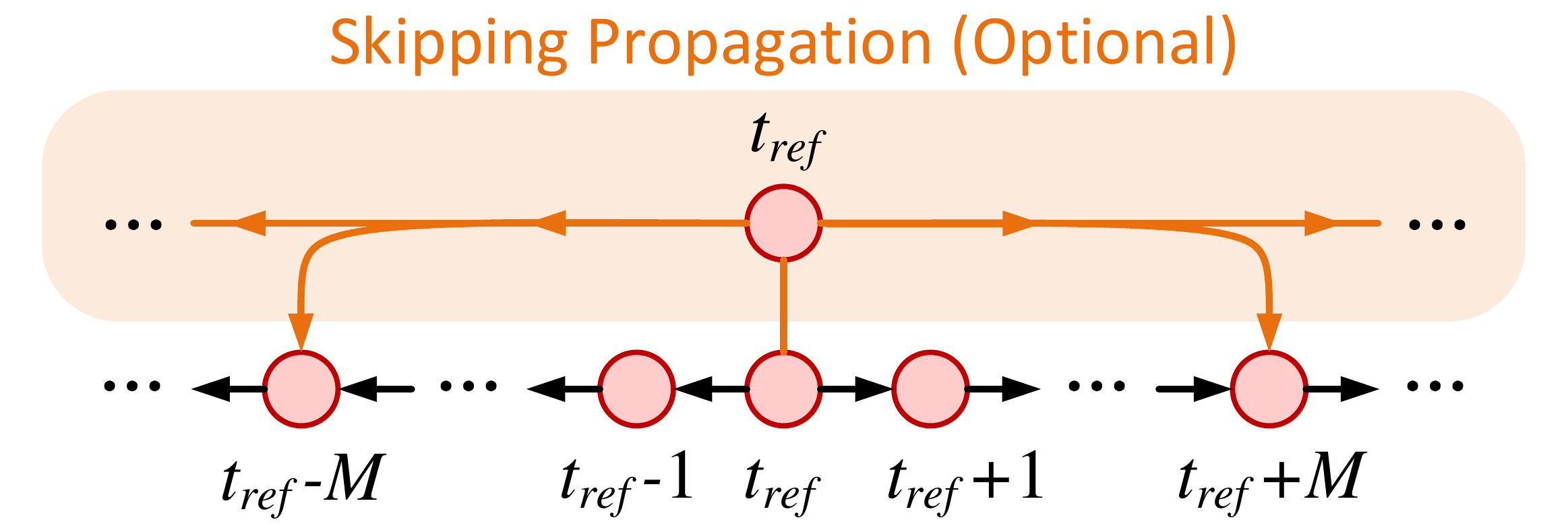}}
	\caption{Propagating the reference key point in the reference frame to other frames for initialization. Skipping propagation is only used for some long input sequences.}
	\label{fig:propagation}
    \vspace{-0.5em}
\end{figure}

\begin{figure*}[t]
    \vspace{-0.7em}
	\centerline{\includegraphics[width=1.0\linewidth]{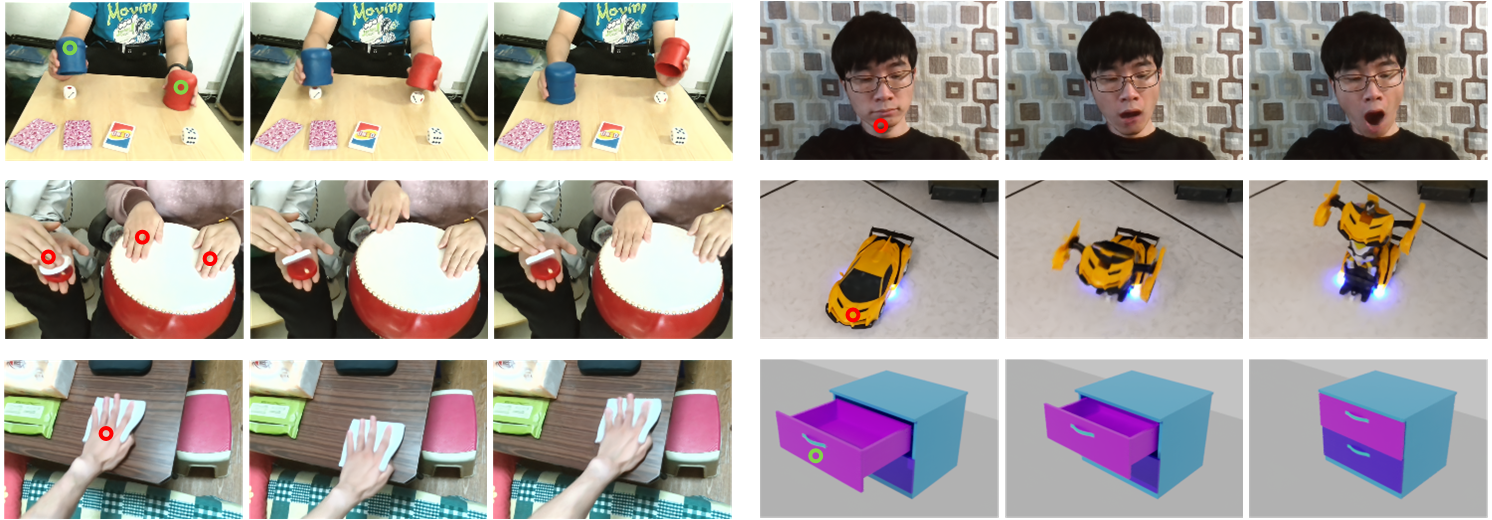}}
    \vspace{-0.5em}
	\caption{Our editing results on various scenes. The first image of each scene also shows the key points (circled in red or green). 
    The \emph{car transformer} sequence on the right side is provided by CoNeRF \cite{kania2022conerf}. The last row on the right side is from a synthetic sequence.}
	\label{fig:results}
    \vspace{-0.5em}
\end{figure*}

Now for each key point, we have a reference key point position and a corresponding reference frame. To initialize key point positions in the whole sequence, we propagate this reference key point to other frames by optical flow from pre-trained RAFT~\cite{teed2020raft}. The reference key point is first projected into the input image of the reference frame to get its 2D position, and the 2D position is propagated frame by frame using optical flow as shown in Fig. \ref{fig:propagation}. Then these 2D positions are projected back into 3D space by depth maps from HyperNeRF. Note that there are accumulative errors in this initialization due to frame-by-frame propagation, while these errors will be eliminated in our training stage.

For some very long input sequences, we found that the initialization method above may not perform well. This is because the accumulative errors may become too large, and the key points in the image space may be propagated into other objects (e.g., background). So we propose a skipping propagation method as shown in Fig. \ref{fig:propagation}. For each key point, we additionally propagate the reference key point in the reference frame every $M$ frames (i.e., $t_{ref}$ to $t_{ref} + M$, $t_{ref}$ to $t_{ref} + 2M$, and so on), and replace the frame-by-frame positions if their confidences are greater than a threshold. This confidence is calculated by the consistency between the forward optical flow and the backward optical flow: 
\begin{equation}
    \label{eq:confidence}
    Conf(t) = \left\| F^{t_{ref}}_{t} (F_{t_{ref}}^{t}(\widehat{k}_{ref})) - \widehat{k}_{ref} \right\|^{-1}, 
\end{equation}
where $t = t_{ref} + i M$, $i \in \mathbb{Z}$, and the hat of $\widehat{k}_{ref}$ indicates that it is a 2D position in the reference frame.

\subsection{Editing by Key Points}
\label{sec:editing}

After training, users can easily edit the modeled scenes by feeding the network with desired key point positions. As the key points are in the 3D space, our method supports up to three-dimensional editing for each part. We also provide a graphical user interface (GUI) in Sec. \ref{sec:applcations}.
\section{Experiments}

We show some results after editing in Fig. \ref{fig:results}. Some effects, like shadow and reflection changes, can also be edited correctly. And please refer to our accompanying video and supplementary materials for more results and experiments.

\subsection{Implementation Details}

We set the weight of motion loss to $10^{-4}$, the weight of geometry loss to $0.5$, and the weight of warp regularization loss to $0.1$. The real data is captured in a resolution of $1280 \times 720$ and down-sampled to $320 \times 180$ for network training. Our network is trained on 4 NVIDIA GeForce RTX 3090 graphics cards, and takes around 5 hours for training with 250k iterations. Our code is based on HyperNeRF~\cite{park2021hypernerf}. And the camera poses of input frames are solved by COLMAP \cite{schonberger2016structure}. More details and discussions of our method are provided in our supplementary materials.

\begin{figure*}[ht]
    \vspace{-1.5em}
	\centerline{\includegraphics[width=1.0\linewidth]{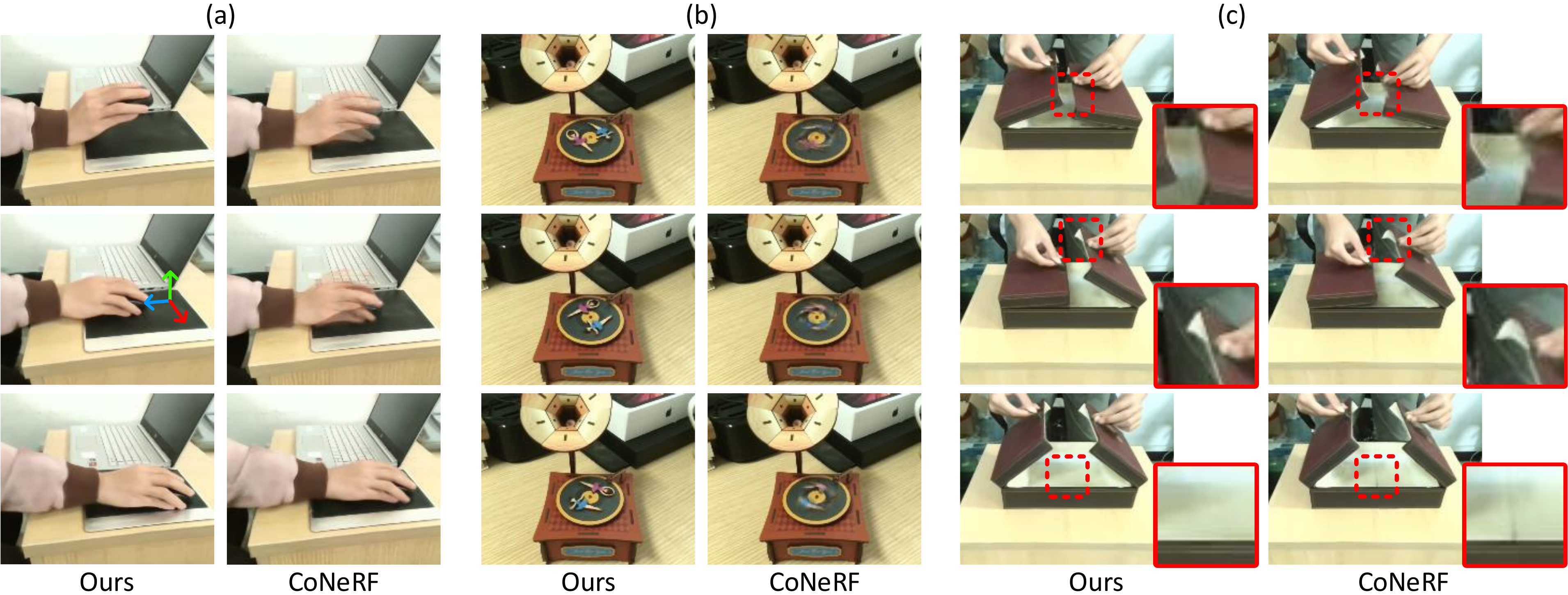}}
	\caption{Qualitative comparisons with CoNeRF~\cite{kania2022conerf}. Our method does not require user annotations for training and supports multi-dimensional editing. Note that the rotations in (b) also cannot be represented by the one-dimensional attribute values in CoNeRF.}
	\label{fig:comp_conerf}
    \vspace{-0.5em}
\end{figure*}

\begin{table*}[ht]
    \centering
    \setlength\tabcolsep{12pt}
    \begin{tabular}{l|ccc|ccc}
        \toprule
        & \multicolumn{3}{|c|}{Reconstruction} & \multicolumn{3}{|c}{Editing} \\
        Method & PSNR $\uparrow$ & MS-SSIM $\uparrow$ & LPIPS $\downarrow$ & PSNR $\uparrow$ & MS-SSIM $\uparrow$ & LPIPS $\downarrow$\\
        \hline
        HyperNeRF \cite{park2021hypernerf} & 42.67 & 0.9964 & 0.0823 & - & - & - \\
        CoNeRF \cite{kania2022conerf} & 40.40 & 0.9891 & 0.0947 & 39.79 & 0.9886 & 0.0949\\
        Ours & \textbf{44.35} & \textbf{0.9973} & \textbf{0.0808} & \textbf{40.01} & \textbf{0.9956} & \textbf{0.0822}\\
        \bottomrule
    \end{tabular}
    \caption{Quantitative comparisons of reconstruction and editing qualities on synthetic data. The reconstruction qualities are measured by the errors in novel-view synthesis. We report PSNR, MS-SSIM \cite{wang2003multiscale}, and LPIPS \cite{zhang2018unreasonable}. Our method performs the best.}
	\label{tab:syn_comp}
\end{table*}

\subsection{Comparisons}
\label{sec:comparisons}

Here we compare our method with state-of-the-art methods HyperNeRF~\cite{park2021hypernerf} and CoNeRF~\cite{kania2022conerf}. HyperNeRF is capable of topologically varying scene reconstruction but does not enable scene editing. CoNeRF allows topologically varying editing but only supports one-dimensional editing for each dynamic part, and user annotations, including masks for every part and their attribute values, are necessary for its pipeline. 
For example, to train a CoNeRF network on an opening mouth sequence, users have to select some input frames, mask the mouth regions in these frames, and set the corresponding attribute values to $1$ when the mouth is open and $-1$ when the mouth is closed. 

\textbf{Qualitative results.} As our method focuses on editing, we first evaluate the editing ability of our method. We compare our method with CoNeRF, while HyperNeRF does not support editing.

We show some editing results of ours and CoNeRF in Fig. \ref{fig:comp_conerf}. Firstly, as shown in the (a) results, our method based on 3D key points enables up to three-dimensional editing, while CoNeRF fails to encode multi-dimensional dynamics by using one-dimensional attribute values. Secondly, our training stage is fully automatic without user annotations. Especially when different parts are close to each other, it becomes quite difficult for end-users to provide very accurate masks at the boundaries, which further leads to artifacts in CoNeRF as shown in (c) of Fig. \ref{fig:comp_conerf}. In contrast, ours can distinguish different parts automatically. Besides, our editing method allows users to drag the key points to their desired positions, which is more intuitive than inputting attribute values as in CoNeRF.

\textbf{Quantitative results on synthetic data.} Here we compare our method with HyperNeRF and CoNeRF on synthetic data. We use three data sequences synthesized by Kubric~\cite{greff2022kubric}, and each contains 400 frames for training. Some results on one of these sequences are shown in Fig. \ref{fig:results}. For CoNeRF training, we annotate $5\%$ frames in the training set using ground truth masks and ground truth attribute values, which are the same as the experiment settings in CoNeRF paper~\cite{kania2022conerf}. While for our method, we still use the optical flow from RAFT and depth maps from HyperNeRF. We \emph{do not} use ground truth optical flow or ground truth depth maps, keeping these settings the same as for real data. Both compared methods are implemented by the original authors and are trained with the same batch size and iteration step as ours.

First, we compare the reconstruction qualities of these methods by rendering the same synthetic scenes in novel viewpoints and evaluating the novel-view synthesis abilities. As shown in Table \ref{tab:syn_comp}, our method reaches the best performance. Note that our method even slightly outperforms HyperNeRF on this task. This is because, in our method, the frames with similar motions are initialized with similar key point positions, while HyperNeRF uses random initialization. Thus, it is easier for our network to integrate the information from the frames with similar motions but in different viewpoints. An experiment on this is further provided in our supplementary materials.


Next, we compare the editing qualities of our method and CoNeRF, while HyperNeRF cannot be used for editing. We derive ground truth key point positions and attribute values from ground truth motions, then use them to edit the scenes in our method and CoNeRF, respectively. Errors are computed between the ground truth images and the rendered images after editing. Our method also outperforms CoNeRF as shown in Table \ref{tab:syn_comp}.

\begin{table}[t]
    \centering
    \begin{tabular}{l|ccc}
        \toprule
        Method & PSNR $\uparrow$ & MS-SSIM $\uparrow$ & LPIPS $\downarrow$ \\
        \hline
        HyperNeRF \cite{park2021hypernerf} & 30.56 & 0.9864 & 0.1281 \\
        CoNeRF \cite{kania2022conerf} & 30.65 & 0.9869 & 0.1307 \\
        Ours & 30.67 & 0.9869 & 0.1314 \\
        \bottomrule
    \end{tabular}
    \caption{Quantitative comparisons of interpolation qualities on real data.}
	\label{tab:real_interp}
\end{table}

\textbf{Quantitative results on real data.} As it is difficult to obtain novel-view ground truth for real data sequences, we turn to compare the interpolation qualities on real data. For a real data sequence with $2N$ frames, we pick out $N$ frames with even indices as the training set, and the other $N$ frames with odd indices are used as ground truth for testing. Table \ref{tab:real_interp} demonstrates that all the compared methods get similar quantitative results on this task. Key point positions for our method, attribute values for CoNeRF, and all the latent codes are interpolated in this task. In CoNeRF training, we select $1\%$ frames with extreme attribute values for annotations, as recommended by CoNeRF \cite{kania2022conerf}.

Besides, when rendering the same real scene with two dynamic parts in the same resolution, CoNeRF takes 1.77s, HyperNeRF takes 0.95s, while ours takes 0.90s. And CoNeRF needs to add a new MLP for each dynamic part, which makes its network not as compact as the other methods.

\begin{table}[t]
    \vspace{-1em}
    \centering
    \small
    \setlength\tabcolsep{5pt}
    \begin{tabular}{l|ccc}
        \toprule
        Method & PSNR $\uparrow$ & MS-SSIM $\uparrow$ & LPIPS $\downarrow$ \\
        \hline
        Base (w/o supervision) & 24.60 & 0.8540 & 0.1680 \\
        + $L_{motion}$ & 29.79 & 0.9429 & 0.1150 \\
        + $L_{motion}$ + $L_{geo}$ & 32.59 & 0.9670 & 0.1013 \\
        + $L_{motion}$ + $L_{geo}$ + init & \textbf{40.01} & \textbf{0.9956} & \textbf{0.0822} \\
        \bottomrule
    \end{tabular}
    \caption{Ablation studies. We evaluate the motion loss, the geometry loss, and the initialization stage. Our final method in the last row performs the best.}
	\label{tab:ablation}
\end{table}

\subsection{Ablation Studies}

Our method makes use of 3D key points with the help of 2D optical flow and 1D depth maps. They are first utilized to initialize key point positions, then to formulate the motion loss and the geometry loss. We evaluate the two losses and the initialization stage in Table \ref{tab:ablation} by measuring the editing qualities on synthetic data. The base method does not use any information from optical flow or depth maps, and its modeled scene changes randomly according to key point movements, while our final method in the last row of Table \ref{tab:ablation} reaches the best performance.


\begin{figure}[t]
	\centerline{\includegraphics[width=1.0\linewidth]{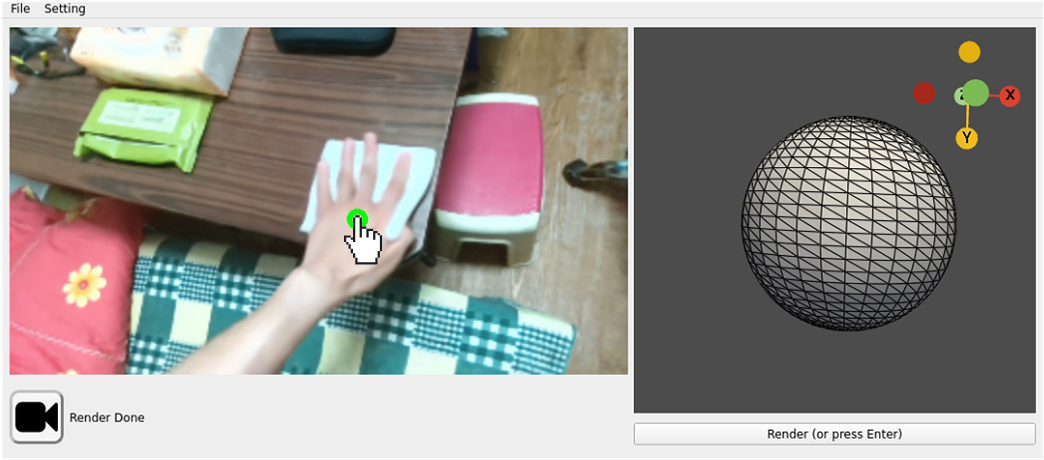}}
	\caption{Graphical user interface. The left widget shows the rendered image and the corresponding draggable key points. 
    The right widget allows the user to change the viewpoint.
    Our GUI also supports drawing a trail of key points, then rendering a video.
    }
	\label{fig:GUI}
\end{figure}

\subsection{Applications}
\label{sec:applcations}

\textbf{Graphical user interface.} We implement a graphical user interface (GUI) for editing and novel-view synthesis, which is shown in Fig. \ref{fig:GUI} and in our accompanying video. Note that end-users actually drag the key points in the 2D interface, so we provide 1D default depth values for key points to form 3D positions, and we also allow end-users to further edit these depth values. The default depth value for a pixel position is obtained by finding $K$ closest key point positions in the input sequence after projecting into the current view and computing their average depth.

\textbf{Novel scenes generation.} Novel scenes that are unseen in the training sequence can also be generated by our method. For example, in the piano toy sequence of Fig. \ref{fig:teaser}, the input data only contains knocking on each piano key, while our method can generate sliding on the piano keys by interpolation. And our method can also combine various dynamics of different parts to create novel scenes, such as the dice cups sequence results shown in Fig. \ref{fig:teaser}.

\textbf{Motion transfer.} Once reconstructed, our modeled scenes can be driven by motions from other sequences. We show a phonograph toy driven by a disk in our accompanying video. 
Optical flow is used to track a manually selected point on the source video and an affine transformation then maps the tracked point into our key point space.

\begin{figure}[t]
    \vspace{-1em}
	\centerline{\includegraphics[width=1.0\linewidth]{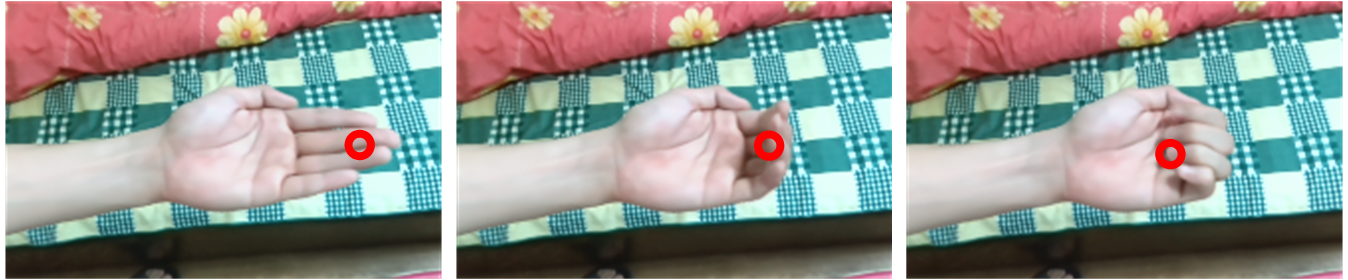}}
	\caption{Editing results on a challenging scene where the selected key point is not always visible in the full sequence. 
    }
	\label{fig:limitation}
\end{figure}

\subsection{Discussions}

We build our framework based on surface key points. While in some challenging sequences, there may not exist a proper surface point that is visible in all frames to become a key point. Our method can still get plausible results on these sequences, but the consistency of key points is not as good as in other scenes, as shown in Fig. \ref{fig:limitation}.

\textbf{Limitations}. We assume that the dynamics of a canonical location mainly depend on one key point. If the scene becomes very complex that does not satisfy this assumption (e.g., a dancing human), our method may fail. Also, it's hard for our method to pick out surface key points for semi-transparent objects like smoke. Extrapolation cannot be performed well for our method when the key points are dragged too far away from their positions in the training sequence. Our method supports multi-dimensional editing, but if the captured objects only have one-dimensional dynamics (e.g., drawer only moves in 1D), our method can only generate one-dimensional dynamics. Besides, our method cannot work well when RAFT or HyperNeRF fails. 

\section{Conclusions}

We propose EditableNeRF, editable topologically varying neural radiance fields that enable end-users to easily edit dynamic scenes. The key to achieving this is to build our framework by leveraging weighted key points to model topologically varying dynamics, which further achieves intuitive multi-dimensional editing. And a scene analysis method that can measure the dynamics in the scene is also proposed to detect and further initialize these key points. Our method is trained fully automatically using a single-view input sequence and can be easily used by end-users, bringing new applications for editable photo-realistic novel-view synthesis.


\noindent \textbf{Acknowledgements.} \small This work was supported by the National Key R\&D Program of China (2018YFA0704000), Beijing Natural Science Foundation (M22024), the NSFC (No.62021002), and the Key Research and Development Project of Tibet Autonomous Region (XZ202101ZY0019G). This work was also supported by THUIBCS, Tsinghua University, and BLBCI, Beijing Municipal Education Commission. Feng Xu is the corresponding author.

{\small
\bibliographystyle{ieee_fullname}
\bibliography{egbib}

\begin{thebibliography}{10}\itemsep=-1pt

\bibitem{attal2021torf}
Benjamin Attal, Eliot Laidlaw, Aaron Gokaslan, Changil Kim, Christian Richardt,
  James Tompkin, and Matthew O'Toole.
\newblock T{\"o}rf: Time-of-flight radiance fields for dynamic scene view
  synthesis.
\newblock {\em Advances in Neural Information Processing Systems},
  34:26289--26301, 2021.

\bibitem{boss2021nerd}
Mark Boss, Raphael Braun, Varun Jampani, Jonathan~T Barron, Ce Liu, and Hendrik
  Lensch.
\newblock Nerd: Neural reflectance decomposition from image collections.
\newblock In {\em Proceedings of the IEEE/CVF International Conference on
  Computer Vision}, pages 12684--12694, 2021.

\bibitem{chen2021animatable}
Jianchuan Chen, Ying Zhang, Di Kang, Xuefei Zhe, Linchao Bao, Xu Jia, and
  Huchuan Lu.
\newblock Animatable neural radiance fields from monocular rgb videos.
\newblock {\em arXiv preprint arXiv:2106.13629}, 2021.

\bibitem{deng2021deformed}
Yu Deng, Jiaolong Yang, and Xin Tong.
\newblock Deformed implicit field: Modeling 3d shapes with learned dense
  correspondence.
\newblock In {\em Proceedings of the IEEE/CVF Conference on Computer Vision and
  Pattern Recognition}, pages 10286--10296, 2021.

\bibitem{dou2017motion2fusion}
Mingsong Dou, Philip Davidson, Sean~Ryan Fanello, Sameh Khamis, Adarsh Kowdle,
  Christoph Rhemann, Vladimir Tankovich, and Shahram Izadi.
\newblock Motion2fusion: Real-time volumetric performance capture.
\newblock {\em ACM Transactions on Graphics (TOG)}, 36(6):1--16, 2017.

\bibitem{du2018montage4d}
Ruofei Du, Ming Chuang, Wayne Chang, Hugues Hoppe, and Amitabh Varshney.
\newblock Montage4d: interactive seamless fusion of multiview video textures.
\newblock In {\em Proceedings of the ACM SIGGRAPH Symposium on Interactive 3D
  Graphics and Games}, pages 1--11, 2018.

\bibitem{flynn2019deepview}
John Flynn, Michael Broxton, Paul Debevec, Matthew DuVall, Graham Fyffe, Ryan
  Overbeck, Noah Snavely, and Richard Tucker.
\newblock Deepview: View synthesis with learned gradient descent.
\newblock In {\em Proceedings of the IEEE/CVF Conference on Computer Vision and
  Pattern Recognition}, pages 2367--2376, 2019.

\bibitem{gafni2021dynamic}
Guy Gafni, Justus Thies, Michael Zollhofer, and Matthias Nie{\ss}ner.
\newblock Dynamic neural radiance fields for monocular 4d facial avatar
  reconstruction.
\newblock In {\em Proceedings of the IEEE/CVF Conference on Computer Vision and
  Pattern Recognition}, pages 8649--8658, 2021.

\bibitem{gao2022nerf}
Kyle Gao, Yina Gao, Hongjie He, Denning Lu, Linlin Xu, and Jonathan Li.
\newblock Nerf: Neural radiance field in 3d vision, a comprehensive review.
\newblock {\em arXiv preprint arXiv:2210.00379}, 2022.

\bibitem{grassal2022neural}
Philip-William Grassal, Malte Prinzler, Titus Leistner, Carsten Rother,
  Matthias Nie{\ss}ner, and Justus Thies.
\newblock Neural head avatars from monocular rgb videos.
\newblock In {\em Proceedings of the IEEE/CVF Conference on Computer Vision and
  Pattern Recognition}, pages 18653--18664, 2022.

\bibitem{greff2022kubric}
Klaus Greff, Francois Belletti, Lucas Beyer, Carl Doersch, Yilun Du, Daniel
  Duckworth, David~J Fleet, Dan Gnanapragasam, Florian Golemo, Charles
  Herrmann, et~al.
\newblock Kubric: A scalable dataset generator.
\newblock In {\em Proceedings of the IEEE/CVF Conference on Computer Vision and
  Pattern Recognition}, pages 3749--3761, 2022.

\bibitem{guo2017real}
Kaiwen Guo, Feng Xu, Tao Yu, Xiaoyang Liu, Qionghai Dai, and Yebin Liu.
\newblock Real-time geometry, albedo, and motion reconstruction using a single
  rgb-d camera.
\newblock {\em ACM Transactions on Graphics (TOG)}, 36(3):1--13, 2017.

\bibitem{guo2021ad}
Yudong Guo, Keyu Chen, Sen Liang, Yong-Jin Liu, Hujun Bao, and Juyong Zhang.
\newblock Ad-nerf: Audio driven neural radiance fields for talking head
  synthesis.
\newblock In {\em Proceedings of the IEEE/CVF International Conference on
  Computer Vision}, pages 5784--5794, 2021.

\bibitem{hong2022headnerf}
Yang Hong, Bo Peng, Haiyao Xiao, Ligang Liu, and Juyong Zhang.
\newblock Headnerf: A real-time nerf-based parametric head model.
\newblock In {\em Proceedings of the IEEE/CVF Conference on Computer Vision and
  Pattern Recognition}, pages 20374--20384, 2022.

\bibitem{jang2021codenerf}
Wonbong Jang and Lourdes Agapito.
\newblock Codenerf: Disentangled neural radiance fields for object categories.
\newblock In {\em Proceedings of the IEEE/CVF International Conference on
  Computer Vision}, pages 12949--12958, 2021.

\bibitem{kania2022conerf}
Kacper Kania, Kwang~Moo Yi, Marek Kowalski, Tomasz Trzci{\'n}ski, and Andrea
  Tagliasacchi.
\newblock Conerf: Controllable neural radiance fields.
\newblock In {\em Proceedings of the IEEE/CVF Conference on Computer Vision and
  Pattern Recognition}, pages 18623--18632, 2022.

\bibitem{li2022neural}
Tianye Li, Mira Slavcheva, Michael Zollhoefer, Simon Green, Christoph Lassner,
  Changil Kim, Tanner Schmidt, Steven Lovegrove, Michael Goesele, Richard
  Newcombe, et~al.
\newblock Neural 3d video synthesis from multi-view video.
\newblock In {\em Proceedings of the IEEE/CVF Conference on Computer Vision and
  Pattern Recognition}, pages 5521--5531, 2022.

\bibitem{li2021neural}
Zhengqi Li, Simon Niklaus, Noah Snavely, and Oliver Wang.
\newblock Neural scene flow fields for space-time view synthesis of dynamic
  scenes.
\newblock In {\em Proceedings of the IEEE/CVF Conference on Computer Vision and
  Pattern Recognition}, pages 6498--6508, 2021.

\bibitem{liu2021neural}
Lingjie Liu, Marc Habermann, Viktor Rudnev, Kripasindhu Sarkar, Jiatao Gu, and
  Christian Theobalt.
\newblock Neural actor: Neural free-view synthesis of human actors with pose
  control.
\newblock {\em ACM Transactions on Graphics (TOG)}, 40(6):1--16, 2021.

\bibitem{liu2021editing}
Steven Liu, Xiuming Zhang, Zhoutong Zhang, Richard Zhang, Jun-Yan Zhu, and
  Bryan Russell.
\newblock Editing conditional radiance fields.
\newblock In {\em Proceedings of the IEEE/CVF International Conference on
  Computer Vision}, pages 5773--5783, 2021.

\bibitem{lombardi2019neural}
Stephen Lombardi, Tomas Simon, Jason Saragih, Gabriel Schwartz, Andreas
  Lehrmann, and Yaser Sheikh.
\newblock Neural volumes: learning dynamic renderable volumes from images.
\newblock {\em ACM Transactions on Graphics (TOG)}, 38(4):1--14, 2019.

\bibitem{loper2015smpl}
Matthew Loper, Naureen Mahmood, Javier Romero, Gerard Pons-Moll, and Michael~J
  Black.
\newblock Smpl: A skinned multi-person linear model.
\newblock {\em ACM Transactions on Graphics (TOG)}, 34(6):1--16, 2015.

\bibitem{mildenhall2020nerf}
B Mildenhall.
\newblock Nerf: Representing scenes as neural radiance fields for view
  synthesis.
\newblock In {\em European Conference on Computer Vision}, pages 405--421.
  Springer, 2020.

\bibitem{mildenhall2019local}
Ben Mildenhall, Pratul~P Srinivasan, Rodrigo Ortiz-Cayon, Nima~Khademi
  Kalantari, Ravi Ramamoorthi, Ren Ng, and Abhishek Kar.
\newblock Local light field fusion: Practical view synthesis with prescriptive
  sampling guidelines.
\newblock {\em ACM Transactions on Graphics (TOG)}, 38(4):1--14, 2019.

\bibitem{noguchi2021neural}
Atsuhiro Noguchi, Xiao Sun, Stephen Lin, and Tatsuya Harada.
\newblock Neural articulated radiance field.
\newblock In {\em Proceedings of the IEEE/CVF International Conference on
  Computer Vision}, pages 5762--5772, 2021.

\bibitem{park2019deepsdf}
Jeong~Joon Park, Peter Florence, Julian Straub, Richard Newcombe, and Steven
  Lovegrove.
\newblock Deepsdf: Learning continuous signed distance functions for shape
  representation.
\newblock In {\em Proceedings of the IEEE/CVF Conference on Computer Vision and
  Pattern Recognition}, pages 165--174, 2019.

\bibitem{park2021nerfies}
Keunhong Park, Utkarsh Sinha, Jonathan~T Barron, Sofien Bouaziz, Dan~B Goldman,
  Steven~M Seitz, and Ricardo Martin-Brualla.
\newblock Nerfies: Deformable neural radiance fields.
\newblock In {\em Proceedings of the IEEE/CVF International Conference on
  Computer Vision}, pages 5865--5874, 2021.

\bibitem{park2021hypernerf}
Keunhong Park, Utkarsh Sinha, Peter Hedman, Jonathan~T Barron, Sofien Bouaziz,
  Dan~B Goldman, Ricardo Martin-Brualla, and Steven~M Seitz.
\newblock Hypernerf: a higher-dimensional representation for topologically
  varying neural radiance fields.
\newblock {\em ACM Transactions on Graphics (TOG)}, 40(6):1--12, 2021.

\bibitem{peng2021animatable}
Sida Peng, Junting Dong, Qianqian Wang, Shangzhan Zhang, Qing Shuai, Xiaowei
  Zhou, and Hujun Bao.
\newblock Animatable neural radiance fields for modeling dynamic human bodies.
\newblock In {\em Proceedings of the IEEE/CVF International Conference on
  Computer Vision}, pages 14314--14323, 2021.

\bibitem{pumarola2021d}
Albert Pumarola, Enric Corona, Gerard Pons-Moll, and Francesc Moreno-Noguer.
\newblock D-nerf: Neural radiance fields for dynamic scenes.
\newblock In {\em Proceedings of the IEEE/CVF Conference on Computer Vision and
  Pattern Recognition}, pages 10318--10327, 2021.

\bibitem{schonberger2016structure}
Johannes~L Schonberger and Jan-Michael Frahm.
\newblock Structure-from-motion revisited.
\newblock In {\em Proceedings of the IEEE Conference on Computer Vision and
  Pattern Recognition}, pages 4104--4113, 2016.

\bibitem{sitzmann2020implicit}
Vincent Sitzmann, Julien Martel, Alexander Bergman, David Lindell, and Gordon
  Wetzstein.
\newblock Implicit neural representations with periodic activation functions.
\newblock {\em Advances in Neural Information Processing Systems},
  33:7462--7473, 2020.

\bibitem{sitzmann2019deepvoxels}
Vincent Sitzmann, Justus Thies, Felix Heide, Matthias Nie{\ss}ner, Gordon
  Wetzstein, and Michael Zollhofer.
\newblock Deepvoxels: Learning persistent 3d feature embeddings.
\newblock In {\em Proceedings of the IEEE/CVF Conference on Computer Vision and
  Pattern Recognition}, pages 2437--2446, 2019.

\bibitem{sitzmann2019scene}
Vincent Sitzmann, Michael Zollh{\"o}fer, and Gordon Wetzstein.
\newblock Scene representation networks: Continuous 3d-structure-aware neural
  scene representations.
\newblock {\em Advances in Neural Information Processing Systems},
  32:1121--1132, 2019.

\bibitem{srinivasan2021nerv}
Pratul~P Srinivasan, Boyang Deng, Xiuming Zhang, Matthew Tancik, Ben
  Mildenhall, and Jonathan~T Barron.
\newblock Nerv: Neural reflectance and visibility fields for relighting and
  view synthesis.
\newblock In {\em Proceedings of the IEEE/CVF Conference on Computer Vision and
  Pattern Recognition}, pages 7495--7504, 2021.

\bibitem{su2021nerf}
Shih-Yang Su, Frank Yu, Michael Zollh{\"o}fer, and Helge Rhodin.
\newblock A-nerf: Articulated neural radiance fields for learning human shape,
  appearance, and pose.
\newblock {\em Advances in Neural Information Processing Systems},
  34:12278--12291, 2021.

\bibitem{sun2022fenerf}
Jingxiang Sun, Xuan Wang, Yong Zhang, Xiaoyu Li, Qi Zhang, Yebin Liu, and Jue
  Wang.
\newblock Fenerf: Face editing in neural radiance fields.
\newblock In {\em Proceedings of the IEEE/CVF Conference on Computer Vision and
  Pattern Recognition}, pages 7672--7682, 2022.

\bibitem{teed2020raft}
Zachary Teed and Jia Deng.
\newblock Raft: Recurrent all-pairs field transforms for optical flow.
\newblock In {\em European Conference on Computer Vision}, pages 402--419.
  Springer, 2020.

\bibitem{thies2019deferred}
Justus Thies, Michael Zollh{\"o}fer, and Matthias Nie{\ss}ner.
\newblock Deferred neural rendering: Image synthesis using neural textures.
\newblock {\em ACM Transactions on Graphics (TOG)}, 38(4):1--12, 2019.

\bibitem{tretschk2021non}
Edgar Tretschk, Ayush Tewari, Vladislav Golyanik, Michael Zollh{\"o}fer,
  Christoph Lassner, and Christian Theobalt.
\newblock Non-rigid neural radiance fields: Reconstruction and novel view
  synthesis of a dynamic scene from monocular video.
\newblock In {\em Proceedings of the IEEE/CVF International Conference on
  Computer Vision}, pages 12959--12970, 2021.

\bibitem{wang2022clip}
Can Wang, Menglei Chai, Mingming He, Dongdong Chen, and Jing Liao.
\newblock Clip-nerf: Text-and-image driven manipulation of neural radiance
  fields.
\newblock In {\em Proceedings of the IEEE/CVF Conference on Computer Vision and
  Pattern Recognition}, pages 3835--3844, 2022.

\bibitem{wang2021one}
Ting-Chun Wang, Arun Mallya, and Ming-Yu Liu.
\newblock One-shot free-view neural talking-head synthesis for video
  conferencing.
\newblock In {\em Proceedings of the IEEE/CVF Conference on Computer Vision and
  Pattern Recognition}, pages 10039--10049, 2021.

\bibitem{wang2003multiscale}
Zhou Wang, Eero~P Simoncelli, and Alan~C Bovik.
\newblock Multiscale structural similarity for image quality assessment.
\newblock In {\em The Thrity-Seventh Asilomar Conference on Signals, Systems \&
  Computers, 2003}, volume~2, pages 1398--1402. Ieee, 2003.

\bibitem{wei2022self}
Fangyin Wei, Rohan Chabra, Lingni Ma, Christoph Lassner, Michael Zollh{\"o}fer,
  Szymon Rusinkiewicz, Chris Sweeney, Richard Newcombe, and Mira Slavcheva.
\newblock Self-supervised neural articulated shape and appearance models.
\newblock In {\em Proceedings of the IEEE/CVF Conference on Computer Vision and
  Pattern Recognition}, pages 15816--15826, 2022.

\bibitem{xian2021space}
Wenqi Xian, Jia-Bin Huang, Johannes Kopf, and Changil Kim.
\newblock Space-time neural irradiance fields for free-viewpoint video.
\newblock In {\em Proceedings of the IEEE/CVF Conference on Computer Vision and
  Pattern Recognition}, pages 9421--9431, 2021.

\bibitem{xiang2021neutex}
Fanbo Xiang, Zexiang Xu, Milos Hasan, Yannick Hold-Geoffroy, Kalyan Sunkavalli,
  and Hao Su.
\newblock Neutex: Neural texture mapping for volumetric neural rendering.
\newblock In {\em Proceedings of the IEEE/CVF Conference on Computer Vision and
  Pattern Recognition}, pages 7119--7128, 2021.

\bibitem{xie2021fig}
Christopher Xie, Keunhong Park, Ricardo Martin-Brualla, and Matthew Brown.
\newblock Fig-nerf: Figure-ground neural radiance fields for 3d object category
  modelling.
\newblock In {\em 2021 International Conference on 3D Vision (3DV)}, pages
  962--971. IEEE, 2021.

\bibitem{yang2021learning}
Bangbang Yang, Yinda Zhang, Yinghao Xu, Yijin Li, Han Zhou, Hujun Bao, Guofeng
  Zhang, and Zhaopeng Cui.
\newblock Learning object-compositional neural radiance field for editable
  scene rendering.
\newblock In {\em Proceedings of the IEEE/CVF International Conference on
  Computer Vision}, pages 13779--13788, 2021.

\bibitem{yu2021unsupervised}
Hong-Xing Yu, Leonidas Guibas, and Jiajun Wu.
\newblock Unsupervised discovery of object radiance fields.
\newblock In {\em International Conference on Learning Representations}, 2021.

\bibitem{yuan2021revisit}
Yu-Jie Yuan, Yu-Kun Lai, Tong Wu, Lin Gao, and Ligang Liu.
\newblock A revisit of shape editing techniques: From the geometric to the
  neural viewpoint.
\newblock {\em Journal of Computer Science and Technology}, 36(3):520--554,
  2021.

\bibitem{yuan2022nerf}
Yu-Jie Yuan, Yang-Tian Sun, Yu-Kun Lai, Yuewen Ma, Rongfei Jia, and Lin Gao.
\newblock Nerf-editing: geometry editing of neural radiance fields.
\newblock In {\em Proceedings of the IEEE/CVF Conference on Computer Vision and
  Pattern Recognition}, pages 18353--18364, 2022.

\bibitem{zhang2021editable}
Jiakai Zhang, Xinhang Liu, Xinyi Ye, Fuqiang Zhao, Yanshun Zhang, Minye Wu,
  Yingliang Zhang, Lan Xu, and Jingyi Yu.
\newblock Editable free-viewpoint video using a layered neural representation.
\newblock {\em ACM Transactions on Graphics (TOG)}, 40(4):1--18, 2021.

\bibitem{zhang2018unreasonable}
Richard Zhang, Phillip Isola, Alexei~A Efros, Eli Shechtman, and Oliver Wang.
\newblock The unreasonable effectiveness of deep features as a perceptual
  metric.
\newblock In {\em Proceedings of the IEEE Conference on Computer Vision and
  Pattern Recognition}, pages 586--595, 2018.

\bibitem{zhang2021nerfactor}
Xiuming Zhang, Pratul~P Srinivasan, Boyang Deng, Paul Debevec, William~T
  Freeman, and Jonathan~T Barron.
\newblock Nerfactor: Neural factorization of shape and reflectance under an
  unknown illumination.
\newblock {\em ACM Transactions on Graphics (TOG)}, 40(6):1--18, 2021.

\bibitem{zheng2021dtexfusion}
Chengwei Zheng and Feng Xu.
\newblock Dtexfusion: Dynamic texture fusion using a consumer rgbd sensor.
\newblock {\em IEEE Transactions on Visualization and Computer Graphics},
  28(10):3365--3375, 2021.

\bibitem{zheng2022avatar}
Yufeng Zheng, Victoria~Fern{\'a}ndez Abrevaya, Marcel~C B{\"u}hler, Xu Chen,
  Michael~J Black, and Otmar Hilliges.
\newblock Im avatar: Implicit morphable head avatars from videos.
\newblock In {\em Proceedings of the IEEE/CVF Conference on Computer Vision and
  Pattern Recognition}, pages 13545--13555, 2022.

\bibitem{zheng2021deep}
Zerong Zheng, Tao Yu, Qionghai Dai, and Yebin Liu.
\newblock Deep implicit templates for 3d shape representation.
\newblock In {\em Proceedings of the IEEE/CVF Conference on Computer Vision and
  Pattern Recognition}, pages 1429--1439, 2021.

\bibitem{zhou2018stereo}
Tinghui Zhou, Richard Tucker, John Flynn, Graham Fyffe, and Noah Snavely.
\newblock Stereo magnification: learning view synthesis using multiplane
  images.
\newblock {\em ACM Transactions on Graphics (TOG)}, 37(4):1--12, 2018.

\end{thebibliography}
}

\end{document}